\documentclass[preprint,12pt,3p]{elsarticle}

\usepackage{amssymb}
\usepackage{amsmath}
\usepackage{xcolor}
\usepackage{hyperref}
\hypersetup{colorlinks=true}
\usepackage[linesnumbered,ruled]{algorithm2e} 

\SetCommentSty{mycommfont}

\iftrue 
\newcommand{\todo}[1]{{\textcolor{red}{[[TODO: {#1}]]}}}
\newcommand{\hll}[1]{{{{#1}}}}
\newcommand{\hl}[1]{{{{#1}}}}
\newcommand{\highlight}[1]{{{{#1}}}}
\newcommand{\commenttext}[1]{{\textcolor{red}{[[{#1}]]}}}
\newcommand{\commentfoot}[1]{\footnote{\textcolor{red}{\emph{Comment: #1}}}}
\newcommand{\topic}[1]{}
\else
\newcommand{\todo}[1]{}
\newcommand{\commenttext}[1]{}
\newcommand{\commentfoot}[1]{}
\newcommand{\topic}[1]{}
\fi

\iffalse

\else 

\fi

\journal{Computer-Aided Design}

\begin{document}

\begin{frontmatter}

\title{\hll{One-Shot Generation of Near-Optimal Topology through Theory-Driven Machine Learning}}

\author[label1]{Ruijin Cang}
\ead{cruijin@asu.edu}
\address[label1]{Mechanical Engineering, Arizona State University, Tempe}

\author[label1]{Hope Yao}
\ead{houpu.Yao@asu.edu}

\author[label1]{Yi Ren\corref{cor1}}
\ead{yiren@asu.edu}
\ead[url]{designinformaticslab.github.io}
\cortext[cor1]{Corresponding author}

\begin{abstract}
We introduce a theory-driven mechanism for learning a neural network model that performs generative topology design in one shot given a problem setting, circumventing the conventional iterative process that computational design tasks usually entail. The proposed mechanism can lead to machines that quickly response to new design requirements based on its knowledge accumulated through past experiences of design generation. Achieving such a mechanism through supervised learning would require an impractically large amount of problem-solution pairs for training, due to the known limitation of deep neural networks in knowledge generalization. 
To this end, we introduce an interaction between a student (the neural network) and a teacher (the optimality conditions underlying topology optimization): The student learns from existing data and is tested on unseen problems. Deviation of the student's solutions from the optimality conditions is quantified, and used \hll{for choosing} new data points to learn from. \hll{We call this learning mechanism ``theory-driven'', as it explicitly uses domain-specific theories to guide the learning, thus distinguishing itself from purely data-driven supervised learning.}
We show through a compliance minimization problem that \hll{the proposed learning mechanism leads to topology generation with near-optimal structural compliance, much improved from standard supervised learning} under the same computational budget. 
\end{abstract}


\end{frontmatter}

\section{Introduction}
This paper is motivated by the observation that experienced human engineers can quickly generate solutions based on accumulated knowledge, while algorithms are only programmed to solve individual problems from scratch, even when the problems are structured similarly. The lack of ability to generalize from experience makes algorithms often too slow to respond to real-world challenges, especially when a stringent time or cost budget is in place. For example, the design of vehicle body-in-white is often done by experienced structure engineers, since topology optimization (TO) on full-scale crash simulation is not yet fast enough to respond to requests from higher-level design tasks, e.g., geometry design with style and aerodynamic considerations, and thus may slow down the entire design process\footnote{The authors learned about this challenge through personal communication with an engineer at a major car manufacturer.}. 

Research exists in developing deep neural network models that \textit{learn} to create structured solutions in a \textit{one-shot} fashion, circumventing the need of iterations (e.g., in solving systems of equations~\cite{CNNFluid2016}, simulating dynamical systems~\cite{chu2017data}, or searching for optimal solutions~\cite{giryes2018tradeoffs,sosnovik2017neural,ulu2016data}).
Learning of such models through data, however, is often criticized to have limited generalization capability, especially when highly nonlinear input-output relations or high-dimensional output spaces exist~\cite{szegedy2013intriguing,Kurakin2016,Huang2017}. In the context of TO, this means that the network may create structures with unreasonably poor physical properties when it responds to new problem settings. More concretely, consider a topology with a tiny crack in one of its trusses. This design would be far from optimal if the goal is to lower compliance, yet standard data-driven learning mechanisms do not prevent this from happening, i.e., they don't know that they don't know (physics). 

Our goal is to create a learning mechanism that knows what it does not know, and thus can self-improve in an effective way. Specifically, we are curious about how physics-based knowledge, e.g., in the forms of dynamical models, theoretical bounds, and optimality conditions, can be directly injected into the learning of networks that perform one-shot solution generation. We call this type of learning mechanisms ``\textit{theory-driven}''. This idea is particularly enchanting as such knowledge widely exists in engineering disciplines and is often differentiable, which is important for gradient-based learning algorithms. 

As an initial attempt in this direction, this paper will focus on TO, and uses a compliance minimization problem~\cite{wu2018infill} as a running example.
To overview, we introduce an interaction between a student (the neural network) and a teacher (the optimality conditions underlying the TO): The student learns from existing data and is tested on unseen problems. Deviation of the student's solutions from the optimality conditions is quantified, and used to choose new data points for the student to learn from.
We use this mechanism to learn a solution generator from a distribution of compliance minimization problems with different load settings. We show that the proposed method has significantly better performance in predicting the optimal topologies for unseen loads, than using a static data collection, under the same computational cost for data acquisition. 


The proposed method will enable learning of solution generators for engineering problems that need to be solved repetitively with variations in their settings, e.g., optimal design (structures, networks, geometries, etc.) as an inner loop of a larger-scale system design, or real-time optimal control with context-dependent parameters. The generated solutions kick-start iterative solvers with good initial guesses and thus shorten the design (solution searching) processes. 

The rest of the paper is structured as follows: In Sec.~\ref{sec:related} we review related work at the intersection of generative design and machine learning, and highlight the new contributions from this paper. We then revisit in Sec.~\ref{sec:statement} the compliance minimization problem, which will be used as a running example throughout the paper, and explain the need for learning one-shot solution generators. Sec.~\ref{sec:alg} introduces the proposed learning mechanism, which will then be validated in Sec.~\ref{sec:case} against two benchmark methods on the running example. Sec.~\ref{sec:discussion} discusses the connection between the proposed method and recent developments in machine learning, and suggests future directions. Sec.~\ref{sec:conclusion} concludes the paper.

\section{Related Work}
\label{sec:related}

Existing research on generative design answer three types of questions: (1) What is the design \textit{representation}? (2) What is the \textit{goodness} measure of a design? And based on these two, (3) how do we \textit{search} for a good design? 
Challenges in answering these questions include \textit{high-dimensional or ill-defined design spaces} such as for topologies~\cite{bendsoe1988generating,guo2018indirect}, material microstructures~\cite{cang2017microstructure,CANG2018212,BOSTANABAD20181}, or complex geometries~\cite{stiny1980introduction,hsiao1997semantic}, \textit{expensive evaluations} of designs and their sensitivities, e.g., due to model nonlinearity~\cite{allaire2004structural,jung2004topology}, coupled materials or physics~\cite{sigmund2001design,wang2004color,zhu2017two}, or subjective goodness measures~\cite{ren2011design,orbay2015deciphering}, or \textit{search inefficiency} due to the absence of sensitivities~\cite{xie1993simple,querin1998evolutionary,rao2011teaching} or the existence of random variables~\cite{kharmanda2004reliability}. 

This paper takes a different angle by focusing on design tasks for which answers to the above three questions exist, yet applying them to real-world design tasks is computationally unaffordable, thus the need for algorithms that learn to \textit{improve} their efficiency through problem solving. Below we review related work, and explain how this paper is similar and different from them.

To start with, the task of improving learning efficiency through experience is known as \textit{meta-learning}~\cite{vilalta2002perspective,hochreiter2001learning} (with close connections to transfer learning~\cite{caruana1995learning,pan2010survey,ganin2016domain} and life-long learning~\cite{thrun1996learning}). 
While there exists a broad range of problem settings within the literature of meta-learning, the setup commonly involves a student and a learning mechanism that specifies how the student updates its way of learning or solving problems. The student outputs a solution for every input problem, for example, the input can be a labeled dataset and the output a classifier that explains the data, or the input an optimization problem and the output an optimal solution. In the former case, the student updates its way of learning a classifier~\cite{ravi2016optimization} or its hypothesis space of classifiers~\cite{vilalta2002perspective}; in the latter, it updates its gradient~\cite{andrychowicz2016learning} or non-gradient~\cite{chen2016learning} search strategies. These updates are governed by the learning mechanism, the design of which is driven by a goodness measure of the student, e.g., the generalization performance of a classifier or the convergence rate of an optimization solver. 

The problem of learning to generate designs can be cast as a meta-learning problem, where the student is the generator that takes in settings of the design problem (e.g., the distribution and magnitude of loads, the material properties, or the boundary conditions for a topology optimization problem) and outputs a design solution, whereas the learning mechanism updates the architecture or parameters of the generator. 

It is worth noting two differences between the proposed method and contemporary meta-learning. 
First, typical mechanisms proposed in meta-learning literature are iterative, e.g., in forms of recurrent neural networks~\cite{ravi2016optimization,andrychowicz2016learning,wang2016learning}. This choice of model is due to the iterative nature of problem solving and the need of memory in decision making during the iteration. In contrary, we model the transition from problem settings to solutions using a feedforward neural network, which is one-shot in nature. We made this choice based on the finding that optimal solutions to a distribution of TO problems often form a continuous manifold (see Sec.~\ref{sec:statement} for details), which suggests that directly learning the manifold through a feedforward network might be achievable, in which case we circumvent the challenges from modeling iterative solvers. 

Secondly, we note that existing meta-learning tasks are often set in contexts where large data acquisition is affordable. This does not hold in our case, since finite element analysis and design sensitivity analysis are costly yet necessary for training the neural network. This requires us to focus on adaptively choosing data points to improve the generator, thus rendering our approach somewhat more similar to active learning~\cite{tong2001support,settles2012active}, where the goal is to improve data efficiency of learning by querying data based on the learned model. Nonetheless, active learning strategies are usually statistics-based. For example, in the context of classification, new data points are chosen based on uncertainty of their predicted labels~\cite{lewis1994heterogeneous} or their predicted contribution to the prediction error of the learned model~\cite{roy2001toward}. These methods, however, are not suitable for our case since we care about the physical optimality of the generated topologies, rather than the pixel-wise matching between the generated topologies and the corresponding true optima (e.g., l2-norm or cross-entropy defined on image differences often used as metrics of prediction error). The method we introduce is thus distinctively different from active learning, as we choose data based on the optimality conditions of TO, which are problem-dependent and theory-driven.  

Lastly, the proposed learning method is aligned with the recent surge of machine learning techniques with integrated physics knowledge. Among this body of work, \cite{wu2016physics} proposes to learn a neural network for predicting intrinsic physical properties of objects with the assistance of a physical simulator that computes object interactions based on the predicted properties. \cite{jonschkowski2017pves} developed an encoder that computes object positions and velocities from images of objects, by enforcing these properties to be compliant with common sense. Similarly, \cite{stewart2017label} proposed a physics-based regularization to learn object trajectories and human movements from videos. Instead of being taught physics, \cite{denil2016learning} demonstrated a grounded way for machines to acquire an intuition of the physical world through reinforcement learning. Our paper is similar to those learning mechanisms with injected physics constraints, while employing adaptive sampling rather than batch-mode training to avoid expensive simulations. \hll{In addition, the use of theory-driven constraints for learning also aligns our method with Q learning, where values are learned to satisfy Bellman's equation.}





\section{Problem Statement}
\label{sec:statement}
\subsection{One-shot solution generator}
We define a one-shot solution generator as a feedforword neural network $\textbf{x} = g(\textbf{s},\boldsymbol{\theta})$ that computes a solution $\textbf{x} \in \mathcal{X}$ (e.g., a topology) given problem settings $\textbf{s} \in \mathcal{S}$ (e.g., loads) and network parameters $\boldsymbol{\theta}$, \hll{where $\mathcal{X}$ is a solution space, and $\mathcal{S}$ is a problem space (e.g., a space of locations and orientations of loads)}. The generator is one-shot in the sense that computing $\textbf{x}$ through $g$ is much less expensive than using an iterative algorithm. We also define the generalization performance of the generator (denoted by $F(\boldsymbol{\theta})$) as the expected performance of its solutions over a distribution of problems specified by a probability density function $p(\textbf{s})$:
\begin{equation}
    F(\boldsymbol{\theta}) = \mathbb{E}_{p(\textbf{s})} f(g(\textbf{s},\boldsymbol{\theta}), \textbf{s}),
\end{equation}
where $f(\textbf{x},\textbf{s})$ measures the performance of $\textbf{x}$ under $\textbf{s}$.

\subsection{The compliance minimization problem}
We now review the mechanical compliance minimization problem introduced in \cite{wu2018infill} to substantiate $f(\textbf{x},\textbf{s})$. In this context, a solution $\textbf{x}$ is a $N$-by-$1$ vector with values between 0 and 1, elements of which control a density vector, denoted as $\boldsymbol{\rho}$, of the corresponding physical elements of a meshed structure through the following relation: 
\begin{equation}
    \rho_e = \frac{\tanh(\beta/2)+\tanh(\beta(\tilde{x}_e-0.5))}{2\tanh(\beta/2)},
    \label{eq:density}
\end{equation}
where $x_e$ and $\rho_e$ are the elements of $\textbf{x}$ and $\boldsymbol{\rho}$, respectively, for $e=1,\cdots,N$. The shape parameter $\beta$ controls the sharpness of the transition to the density $\rho_e$ from a filtered variable $\tilde{x_e}$, which is a weighted average of neighbours of $x_e$: 
\begin{equation}
    \tilde{x}_e = \frac{\sum_{i\in \mathcal{M}_e} \omega_{i,e} x_i}{\sum_{i \in \mathcal{M}_e} \omega_{i,e}},
    \label{eq:filter1}
\end{equation}
where $\mathcal{M}_e$ is the set of neighbours of element $e$, and weights $\omega_{i,e}$ are defined as 
\begin{equation}
    \omega_{i,e} = 1-||z_i - z_e||_2/r_e,
\end{equation} 
with $z_i$ the coordinates of meshed element $i$, and $r_e$ the filter radius. In TO, this filter (Eq.~\eqref{eq:filter1}) is used to prevent convergence to impractical checkerboard topologies~\cite{wu2018infill}. 

The connection from $\textbf{x}$ to the global stiffness matrix of the topology (denoted as $\textbf{K}$) can be established through the density vector $\boldsymbol{\rho}$. Given loads $\textbf{s}$ and boundary conditions, the displacement $\textbf{u}$ of the structure can be found by solving $\textbf{Ku}=\textbf{s}$, under the assumption that $\textbf{K}$ is independent of $\textbf{u}$ (e.g., linear elastic material and small displacement). The compliance minimization problem can be formulated as 
\begin{equation}
    \begin{aligned}
        \min_{\textbf{x}} & \quad f(\textbf{x},\textbf{s}) = \frac{1}{2}\textbf{u}^T\textbf{K}\textbf{u} \\
        \text{subject to} & \quad \textbf{Ku} = \textbf{s}, \\
        & \quad x_e \in [0, 1], ~\forall e \in \{1,\cdots,N\}, \\
        & \quad g_0(\textbf{x}) = \frac{1}{N}\sum_e \rho_e - \alpha \leq 0,\\
        & \quad g_1(\textbf{x}) = ||\bar{\rho}||_p - \alpha \leq 0.
    \end{aligned}
\label{eq:top}\tag{TO}
\end{equation}
Here the constraint $g_0$ ($g_1$) limits the global (local) density of the structure to be lower than a threshold $\alpha$. $||\textbf{x}||_p = (\frac{1}{N}\sum_e x_e^p)^{1/p}$ is the p-norm defined on $\mathbb{R}^N$. $p$ is set to 16 following \cite{wu2018infill} so that $||\textbf{x}||_p$ approximately computes the maximum of $|\textbf{x}|$ while being differentiable. The averaged local density $\bar{\rho}_e = (\sum_{i \in \mathcal{N}_e} \rho_i)/(\sum_{i \in \mathcal{N}_e} 1)$ is defined on the neighborhood $\mathcal{N}_e = \{i |~ ||x_i - x_e||_2 \leq R_e \}$ with radius $R_e$. Note that $R_e$ for the local density constraint is different from the filter radius $r_e$. The optimality conditions of \eqref{eq:top} are listed as follows
\begin{equation}
    \begin{aligned}
        & \nabla_{\textbf{x}} L := -\textbf{u}^T \frac{\partial \textbf{K}}{\partial \textbf{x}}\textbf{u} + \mu_0 \nabla_{\textbf{x}} g_0 + \mu_1 \nabla_{\textbf{x}} g_1 +
        \boldsymbol{\mu}_u^T - \boldsymbol{\mu}_l^T = \textbf{0}^T  \\
        & x_e \in [0, 1], ~\forall e \in \{1,\cdots,N\}, \\
        & g_0 \leq 0, ~ g_1 \leq 0, ~\mu_0 g_0=0, ~\mu_1 g_1=0, \\
        & \boldsymbol{\mu}_l^T\textbf{x} = 0,
        \boldsymbol{\mu}_u^T(\textbf{x} -\textbf{1}) = 0\\
        & \mu_0 \geq 0, ~ \mu_1 \geq 0, 
        ~\boldsymbol{\mu}_u \geq \textbf{0}, ~\boldsymbol{\mu}_l \geq \textbf{0},
    \end{aligned}
\label{eq:topkkt}
\end{equation}
where $\nabla_{\textbf{x}} y(\textbf{x},\cdot)$ is the partial derivative of function $y$ with respect to variables $\textbf{x}$, and is defined as a row vector. Finding a solution to comply with Eq.~\eqref{eq:topkkt} can be done through a gradient-based solver, e.g., an augmented Lagrangian algorithm (see Sec.~\ref{sec:al}). However, we need to note that the computational cost for converging to an optimal solution usually does not scale well. In particular, solving Eq.~\eqref{eq:topkkt} with a problem size $N=4800$ requires on average around 5000 finite element analyses (i.e., computing $\textbf{u}$ from $\textbf{Ku}=\textbf{s}$).

\subsection{Learning a solution generator}
With the above setup, the problem of learning a one-shot solution generator can be formulated as follows:
\begin{equation}
    \begin{aligned}
        \min_{\boldsymbol{\theta}} & \quad F(\boldsymbol{\theta}) = \mathbb{E}_{p(\textbf{s})} f(\textbf{x},\textbf{s}) \\
        \text{subject to} & \quad \textbf{x} = g(\textbf{s},\boldsymbol{\theta}), \\
        & \quad \textbf{K}\textbf{u} = \textbf{s}, \\
        & \quad f(\textbf{x},\textbf{s}) = \frac{1}{2}\textbf{u}^T\textbf{K}\textbf{u} \\
        & \quad g_0(\textbf{x}) = \frac{1}{N}\sum_e \rho_e - \alpha \leq 0,\\
        & \quad g_1(\textbf{x}) = ||\bar{\rho}||_p - \alpha \leq 0.
    \end{aligned}
    \label{eq:prob}\tag{P}
\end{equation}
We will force network outputs to be within $(0,1)^N$ by attaching sigmoid activations to its output layer.

\section{Learning with a Physics-based Criterion}
\label{sec:alg}
\eqref{eq:prob} can be solved by matching the input-output pairs of the generator to a dataset $\mathcal{D} = \{\textbf{x}_i^{D}, \textbf{s}_i^{D}\}_{i=1}^n$ where $\textbf{x}_i^{D}$ are optimal for $\textbf{s}_i^{D}$. This leads to the data-driven learning formulated as follows:
\begin{equation}
    \begin{aligned}
        \min_{\boldsymbol{\theta}} & \sum_{\mathcal{D}} ||g(\textbf{s}_i^{D},\boldsymbol{\theta}) - \textbf{x}_i^{D}||_2^2.
    \end{aligned}
    \label{eq:fit}\tag{P1}
\end{equation}
As we reviewed in the last section, collecting $\mathcal{D}$ can be costly due to the iterative nature of solving the topology optimization problem \eqref{eq:top}. On the other hand, checking the compliance of an arbitrary topology $\textbf{x}$ to the optimality conditions (Eq.~\eqref{eq:topkkt}) only requires solving $\textbf{Ku}=\textbf{s}$ once. This finding indicates that the optimality conditions may offer affordable means to identify new data point that will most effectively improve the generalization performance of $g$. Specifically, we define the deviation of solution $\textbf{x}$ from the optimality conditions as
\begin{equation}
    d(\textbf{x},\boldsymbol{\mu}) = ||\nabla_{\textbf{x}} L||_2^2 + w_0 g_0^2 + w_1 g_1^2
    \label{eq:deviation}
\end{equation}
where 
the algorithmic parameters $w_0$ and $w_1$ weight the penalties on constraints $g_0$ and $g_1$, respectively. 
One issue in evaluating $d$ is that we do not know the values of the Lagrangian multipliers (which are denoted by $\boldsymbol{\mu}^T = [\mu_0, \mu_1, \boldsymbol{\mu}_l^T, \boldsymbol{\mu}_u^T]$) before solving the problem. To this end, we propose to find $\boldsymbol{\mu}^*$ that minimizes the deviation of $\nabla_{\textbf{x}} L$ from $\textbf{0}$ subject to their constraints from \eqref{eq:prob}:
\begin{equation}
    \begin{aligned}
        \min_{\boldsymbol{\mu}} & \quad ||\nabla_{\textbf{x}} L||_2^2 \\
        \text{subject to} & \quad \boldsymbol{\mu} \geq 0, ~\boldsymbol{\mu}^T \textbf{g}=0.
    \end{aligned}
    \label{eq:mindeviation}\tag{P2}
\end{equation}
By solving \eqref{eq:mindeviation} for all $\textbf{s}$ in a validation set $\mathcal{S}_v$, we can then choose a new training data point $(\textbf{s}^*, \textbf{x}^*)$ for which the minimal deviation $d(g(\textbf{s}^*,\boldsymbol{\theta}),\boldsymbol{\mu}^*)$ is the largest among $\mathcal{S}_v$. It is worth noting that \eqref{eq:mindeviation} is a ($2N+2$)-dimensional quadratic programming problem and can be solved efficiently using standard solvers (e.g., sequential quadratic programming). 

The learning algorithm can now be summarized in Alg.~\ref{alg:learning}. Details on setting the initial training set $\mathcal{S}_0$, the validation set $\mathcal{S}_v$, the computational budget $B$, and the budget lower bound $b$ will be introduced along the case studies in Sec.~\ref{sec:case}.

\begin{algorithm}
\label{alg:learning}
\SetKwInOut{Input}{input}
\SetKwInOut{Output}{output}
\Input{Problem distribution $p(\textbf{s})$}
\Output{Learned model parameters $\boldsymbol{\theta}^*$}
Draw initial problem set $\mathcal{S}_0$ from $p(\textbf{s})$; \par
Find optimal $\textbf{x}_i$ for each $\textbf{s}_i$ in $\mathcal{S}_0$; \par
Initialize dataset $\mathcal{D} = \mathcal{D}_0 = \{(\textbf{x}_i, \textbf{s}_i)\}_{i=1}^{|\mathcal{S}_0|}$; \par
Draw validation problems $\mathcal{S}_v$ from $p(\textbf{s})$; \par
Set computation budget $B = B_0$ and budget lower bound $b$; \par
\While{$B > b$}{
        Update $\boldsymbol{\theta}^*$ by solving \eqref{eq:fit} based on $\mathcal{D}$; \par
        Calculate $\boldsymbol{\mu}_i^*$ and $d(g(\textbf{s}_i,\boldsymbol{\theta}^*),\boldsymbol{\mu}_i^*)$ for all $\textbf{s}_i$ in $\mathcal{S}_v$ based on Eq.~\eqref{eq:deviation}; \par
        Find $\textbf{s}^*\in \mathcal{S}_v$ with the highest $d$ value;\par
        Derive $\textbf{x}^*$ for $\textbf{s}^*$ by solving \eqref{eq:top};\par
        Record $\delta B$ as the number of $\textbf{Ku}=\textbf{s}$ solved in solving \eqref{eq:top} and computing Eq.~\eqref{eq:deviation};\par
        Update the budget $B = B - \delta B$;\par
        $\mathcal{D} = \mathcal{D} + (\textbf{x}^*,\textbf{s}^*)$, $\mathcal{S}_v = \mathcal{S}_v - \textbf{s}^*$;\par
    }
\caption{Theory-driven learning}
\end{algorithm}

\section{Case Studies}
\label{sec:case}
This section presents two case studies where we demonstrate the superior learning efficiency of the proposed algorithm in comparison with two benchmark mechanisms. The first benchmark uses a static dataset ($\mathcal{D}_{\text{static}}$) for training, and is denoted as \textit{Benchmark I}. \highlight{The second benchmark, denoted as \textit{Benchmark II}, is similar to the proposed approach, but chooses data points using a different heuristic.} For all three learning mechanisms \highlight{(Benchmarks I, II, and ours)}, the topology optimization problem is solved by an Augmented Lagrangian algorithm, details of which is deferred to \highlight{the appendix}.

\subsection{The heuristic \highlight{of Benchmark II}}
\highlight{In Benchmark II, we} evaluate the performance of the generator $g$ by measuring the difference between the compliance produced by $g$ and the predicted compliance based on the training data $\mathcal{D}$:
\begin{equation}
    d_h(g(\textbf{s}_i,\boldsymbol{\theta}^*)) = |f(g(\textbf{s}_i,\boldsymbol{\theta}^*),\textbf{s}_i) - \hat{f}(\textbf{s}_i)|,
\end{equation}
where $\textbf{s}_i$ comes from the validation set $\mathcal{S}_v$, and $\hat{f}$ is an ordinary least square model that fits to $\{(\textbf{s}^*_i, f(\textbf{x}^*_i,\textbf{s}^*_i) ~|~ (\textbf{x}^*_i,\textbf{s}^*_i) \in \mathcal{D} \}$. In this study, polynomial models are used for curve fitting. 
The validation data point with the highest value of $d_h$ is chosen, and its true \highlight{optimal} topology is then computed and used to improve the generator.

\subsection{Study setups}
\begin{figure}[htp]
\centering
\includegraphics[width=0.8\textwidth]{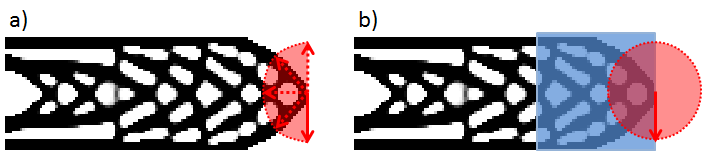}
\caption{Sample solutions for (a) Case 1, where a point load is applied to the middle node at the tip of the cantilever beam, with directions uniformly distributed in $[0,\pi]$, and (b) Case 2, where the point load is applied to a node in the highlighted square area which occupies one-third of the design space, with its direction drawn from $[0,2\pi]$}
\label{fig:top}
\end{figure}

The topology optimization problems to be solved follow \eqref{eq:top}. Two cases are created to demonstrate the scalability of the proposed method. In Case 1, the input $\textbf{s}$ represents a single load applied to a 2D structure represented by a 40-by-120 mesh, see Fig.~\ref{fig:top}a. In this case, the input to the generator is the angle of the load, uniformly distributed in between $0$ and $\pi$, i.e., $p(\textbf{s}) = 1/\pi$. In Case 2, $\textbf{s}$ \highlight{encodes} (1) the x- and y-coordinates of the loading, which is drawn uniformly from all nodes in the highlighted area in Fig.~\ref{fig:top}b, and (2) the direction of the point load uniformly drawn from $0$ to $2\pi$. 

To compare the three learning mechanisms, each is executed 10 times to account for the randomness in the sampling of the initial dataset $\mathcal{D}_0$ (and $\mathcal{D}_{\text{static}}$ in the case of Benchmark I) and the validation set at each iteration. The generalization performance is measured by 
\begin{equation}
    F(\boldsymbol{\theta}^*) = \sum_{\textbf{s}_i \in \mathcal{S}_t} f(g(\textbf{s}_i,\boldsymbol{\theta}^*),\textbf{s}_i),
\end{equation}
where $\mathcal{S}_t$ is a separate test set drawn from $p(\textbf{s})$. For Case 1, the sample sizes are $|\mathcal{S}_0| = 5$, $|\mathcal{S}_v| = 100$, $|\mathcal{S}_t| = 100$, and $|\mathcal{D}_{\text{static}}| = 16$. \highlight{For Case 2, since the problem space is much larger, the sample sizes are set to $|\mathcal{S}_0| = 1000$, $|\mathcal{S}_v| = 6000$, $|\mathcal{S}_t| = 1000$, and $|\mathcal{D}_{\text{static}}| = 7000$. In addition, since validating a large number of inputs becomes expensive, we uniformly sample 100 validation points from $|\mathcal{S}_v|$ to perform active learning in each iteration.} All previously sampled points will be removed from $\mathcal{S}_v$. For both cases, the computational budget $B_0$ is set as $b_{\text{min}}|\mathcal{D}_{\text{static}}|$, where $b_{\text{min}}$ is the minimal solution cost among all problems sampled for Benchmark I. The budget lower bound $b$ is set to the maximal cost among the same set of problems. This setting ensures that the adaptive methods (Benchmark II and the proposed method) will always use less computational resource than the static method for topology optimization, thus creating a comparison in favor of the latter. For the following results, we set $w_0=w_1=1$ for the proposed method. A full parametric study on these hyper-parameters has not been conducted, yet the effectiveness of the current setting is validated (Sec.~\ref{sec:results}). 

\begin{figure}[htp]
\centering
\includegraphics[width=0.4\textwidth]{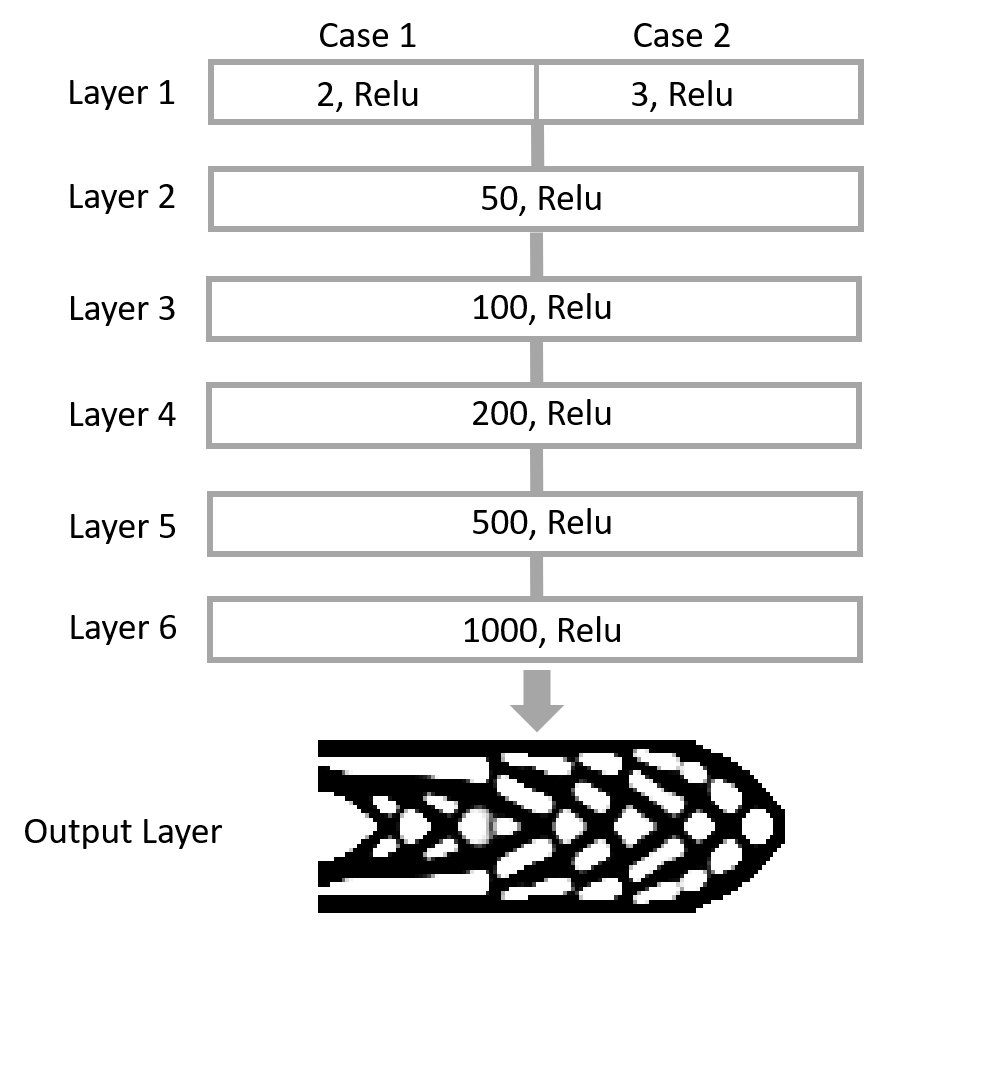}
\caption{Architectures of the solution generators}
\label{fig:generator}
\end{figure}

\highlight{
\subsection{Architectures of the solution generators}
The architectures of the solution generators are summarized in Fig.~\ref{fig:generator}. For Case 1, we use a two-dimensional input to represent the x- and y-components of the point load. For Case 2, \hl{we use a three-dimensional input} to represent the x, y location and the orientation of the point load. The choice of these input representations are based on empirical tests of the generalization performance of the learned models.} \hll{In particular, we found in Case 2 that using a sparse load representation as network inputs will lead to much worse generation performance for all learning mechanisms.}



\subsection{Results}
\label{sec:results}

\begin{figure}[htp]
\centering
\includegraphics[width=1.0\textwidth]{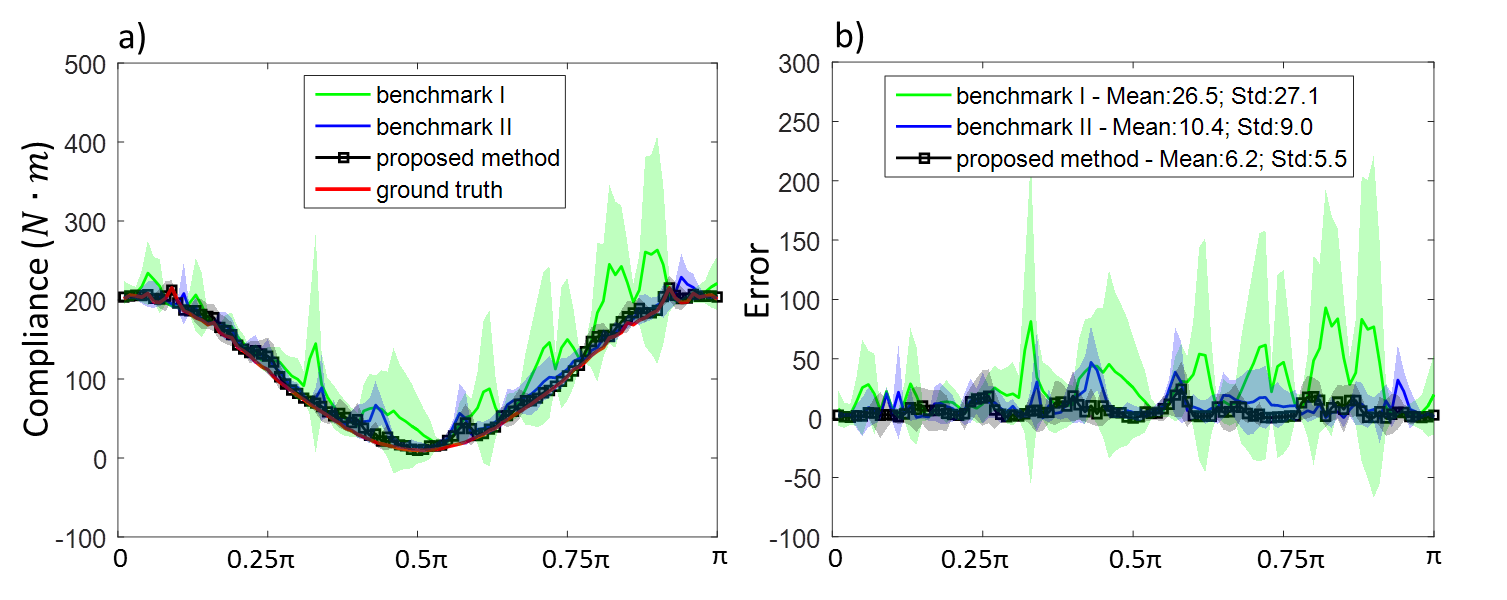}
\caption{(a) Comparison between the compliance of the ground truth topologies and those from the generated solutions from all three learning mechanisms. 10 independent experiments are conducted for each learning mechanism. The means and the standard deviations are reported. (b) The absolute compliance gaps.}
\label{fig:com_results}
\end{figure}

\paragraph{Case 1 results} For Case 1, Fig.~\ref{fig:com_results}a compares the compliance of the topologies generated by all three learning mechanisms with the ground truth for all test inputs; Fig.~\ref{fig:com_results}b reports the corresponding compliance gaps produced by these mechanisms. The generalization performance of a learning mechanism is measured by the mean and the standard deviation of the average compliance gaps across all test inputs. 
The result shows that the proposed mechanism outperforms the benchmarks at predicting optimal topologies for unseen loading conditions in a low-dimensional case. To further demonstrate the difference between the three mechanisms, we visualize and compare generations for four test loads in Fig.~\ref{fig:top}. The loading directions are marked in the last row of the figure, which are $0.3\pi$, $0.4\pi$, $0.45\pi$,$0.55\pi$ respectively. The resultant compliance values are shown at the bottom of each topology. 

\begin{figure}[htp]
\centering
\includegraphics[width=1.0\textwidth]{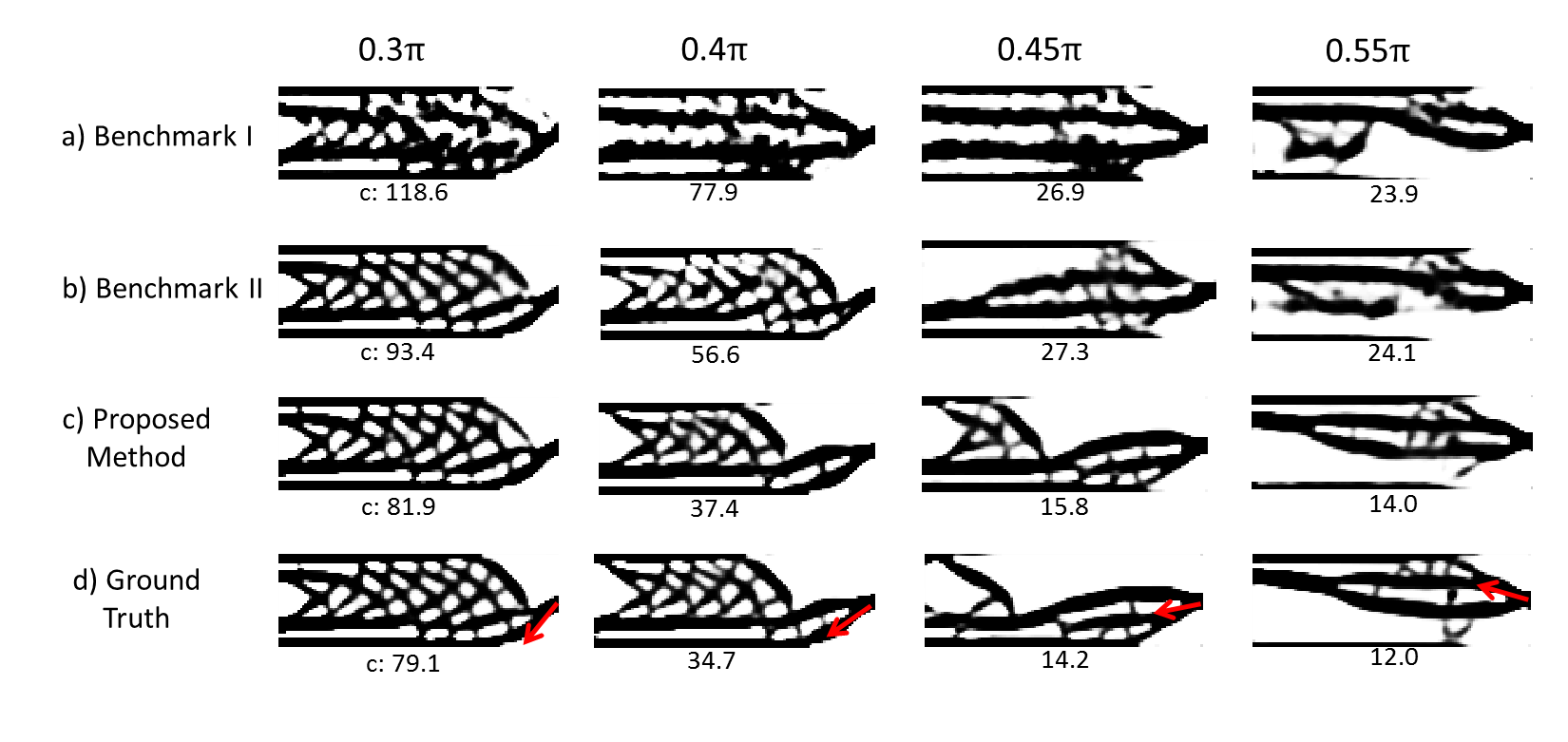}
\caption{Case 1 topologies predicted by (a) Benchmark I, (b) Benchmark II, (c) the proposed method. (d) Ground truth computed using the Augmented Lagrangian solver. Compliance values are shown at the bottom of each topology. Loading directions are marked as the arrows.}
\label{fig:top_com}
\end{figure}

\begin{figure}[htp]
\centering
\includegraphics[width=1.0\textwidth]{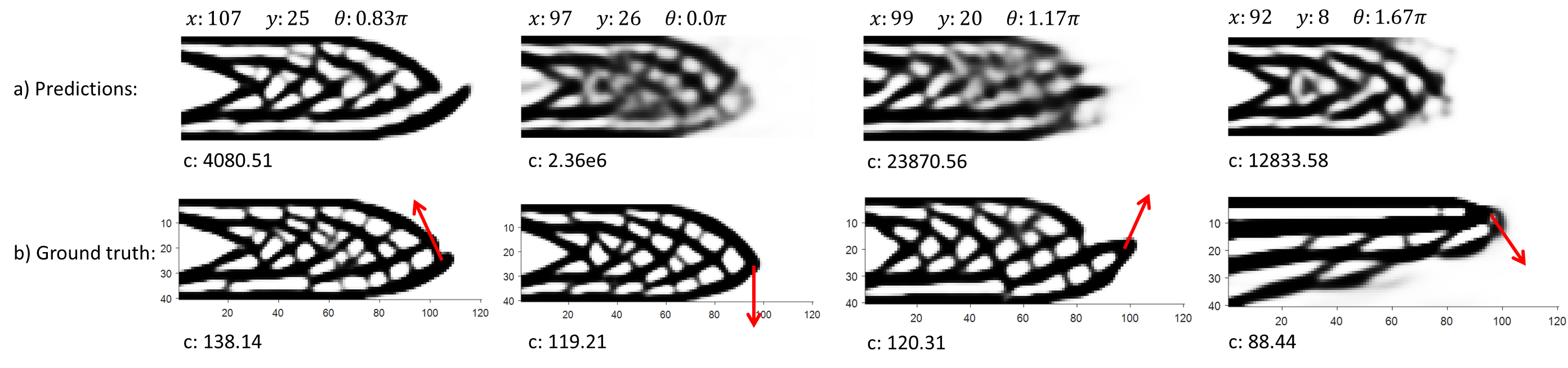}
\caption{Example of failed generations with significant larger compliance than the corresponding ground truth. Loading conditions are marked at the top and compliance at the bottom. Some generations are close to the ground truth in the pixel space but miss critical elements, such as the first three. Some of the generations are simply off, such as the last one.}
\label{fig:defective_design}
\end{figure}

\paragraph{Challenges in Case 2}
\highlight{Case 2 examines the performance of the proposed mechanism under a higher-dimensional and larger input space. We notice that the learning becomes significantly more challenging in this case. see Sec~\ref{sec:discussion}). Specifically, for all three algorithms under the same budget (which is equivalent to solving 7000 TO problems), there exist inputs for which the generated topologies have significantly larger compliance than the ground truth. We mark test data points with a compliance gap of over 1000 as failed designs. 
For Benchmark I, II and the proposed method, the mean failure rates over the entire test are \hl{14$\%$,  0.8$\%$, and 0.6$\%$} respectively.
Some failed generations are shown in Fig.~\ref{fig:defective_design} along side the corresponding ground truth.}

\begin{figure}[htp]
\centering
\includegraphics[width=1.0\textwidth]{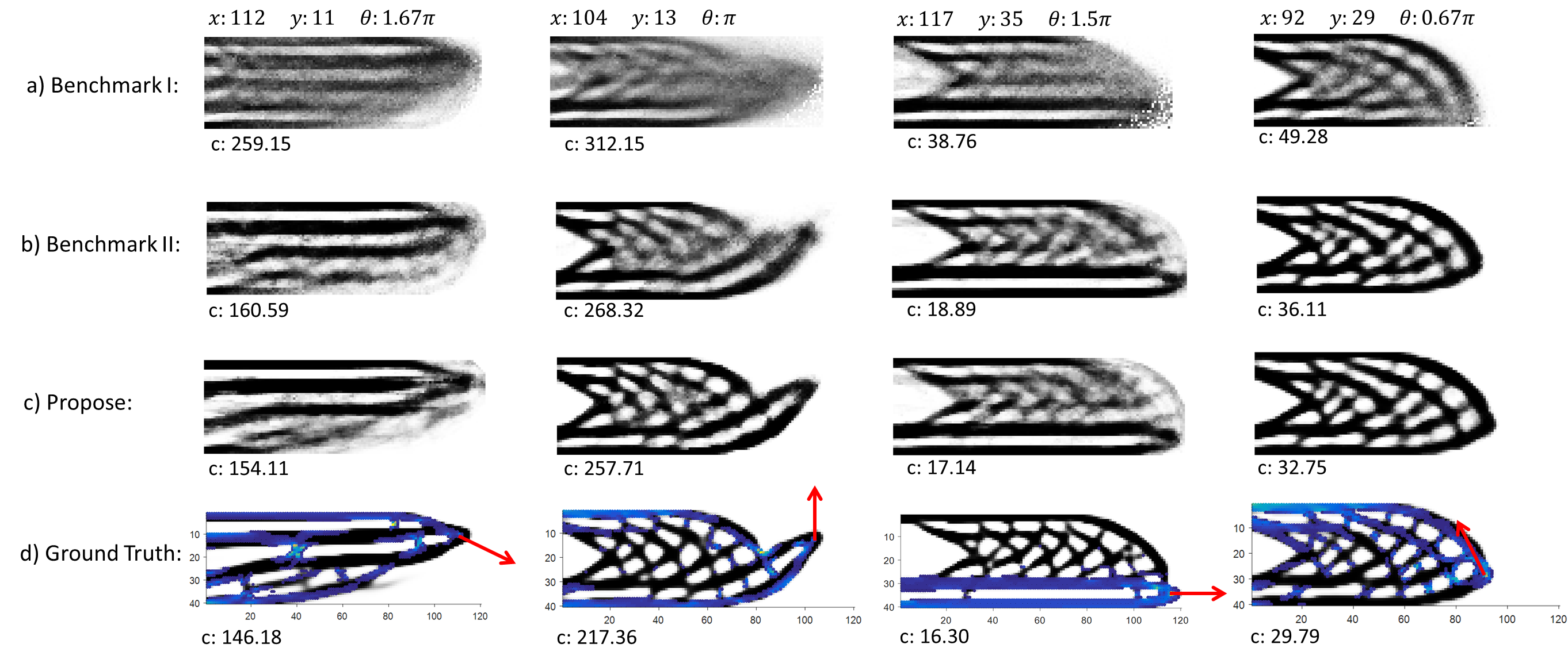}
\caption{\hl{Case 2 optimal topologies predicted by (a) Benchmark I, (b) Benchmark II, and (c) the proposed method. (d) The ground truth topologies with design sensitivity information: cold (warm) color indicates lower (higher) design sensitivity. The point loads are marked as arrows with detailed input values reported at the top of the figure. The compliance values are shown at the bottom of each topology. Best viewed online.}}
\label{fig:top_com_high_dim}
\end{figure}

\highlight{
\paragraph{An augmented learning objective}
To alleviate this issue, we introduce design sensitivity $\nabla_{\textbf{x}} f$ at the optimal solution as a weighting factor of the learning loss in \eqref{eq:fit}, based on the insight that structural elements with higher sensitivity contribute more to the compliance and thus should have higher priority during model training. Concretely, the training problem at each iteration of the active learning process is now formulated as
\begin{equation}
    \begin{aligned}
        \min_{\boldsymbol{\theta}} & \sum_{(\textbf{s}_i, \textbf{x}_i) \in \mathcal{D}}  ||g(\textbf{s}_i,\boldsymbol{\theta}) - \textbf{x}_i||_{\boldsymbol{\Lambda}_i}^2,
    \end{aligned}
    \label{eq:fit2}
\end{equation}
where $\boldsymbol{\Lambda}_i = diag([\lambda_{i,1},\cdots,\lambda_{i,N}])$ is a diagonal weighting matrix, $\lambda_{i,e} = \frac{\nabla_{\textbf{x}} f - \min(\nabla_{\textbf{x}} f)}{\max(\nabla_{\textbf{x}} f)-\min(\nabla_{\textbf{x}} f)}$, and $||\textbf{x}||_{\boldsymbol{\Lambda}}^2 = \textbf{x}^T\boldsymbol{\Lambda}\textbf{x}$. It is important to note that $\boldsymbol{\Lambda}_i$ is a byproduct of computing $\textbf{x}_i$ and does not cost extra budget.}
\highlight{
\paragraph{Case 2 results}
By introducing this augmentation, the mean rate of failures drops to 5.12$\%$, 0.64$\%$, and 0.16$\%$ for Benchmark I, II, and the proposed method, respectively. Since the compliance distributions are far from normal due to the failed designs, we report the means and standard deviations of the median compliance gaps from all experiments and learning mechanisms instead. These results are summarized in Table~\ref{table:com_results_high_dim}. As a demonstration, we compare in Fig.~\ref{fig:top_com_high_dim} the generations from all mechanisms under four test settings. The proposed method has the closest compliance to the ground truth. } 

\begin{table}[]
\centering
\caption{Comparisons on the mean count of failures from 1000 test data points and the median compliance gaps for Case 2}
\begin{tabular}{|l|c|c|}
\hline
               & Failure rate: mean (std) & Median compliance gap: mean (std) \\
\hline
Benchmark I    & \hl{5.12\% (1.2\%)} & 22.12 (3.34)\\
\hline
Benchmark II   & \hl{0.64\% (0.42\%)} & 6.64 (0.87)\\
\hline
Propose method & \hl{\textbf{0.16\%} (0.09\%)} & \textbf{5.2} (0.6)\\
\hline
\end{tabular}
\label{table:com_results_high_dim}
\end{table}

\paragraph{Validity of hyper-parameter settings}
For both case studies we set the weights for the local and global constraints ($w_0$ and $w_1$ in Eq.~\ref{eq:deviation}) to be $w_0 = 1$ and $w_1 = 1$. To validate this setting, we monitor the violation to the constraints by solutions generated for all test inputs. For a proper measure of constraint satisfaction, test cases where constraints are satisfied will be ignored in the calculation of the mean violation. It is noted that ground truth topologies have none-zero violation due to the non-zero error thresholds set in the augmented Lagrangian algorithm. In Case 1 (Case 2), the mean violation to the global volume fraction constraint is $g_0=4.89\%$ (\hl{$3.0\%$}) for the ground truth and $g_0=8.76\%$ (\hl{$1.6\%$}) for the generated solutions. In both cases, no violation to the local constraints are observed for either the ground truth or the generated solutions. This result indicates that the setting of the hyper-parameters leads to the learning of a generator with reasonable compliance to the global and local volume fraction constraints. A parametric study is yet needed to fully characterize the tradeoff between the learning of effective topologies and that of constraint compliance by tuning the hyper-parameters.

\highlight{
\paragraph{Comparison on solution generation speed} Lastly, we shall note that the computation time required for generating solutions through the learned model is negligible (in the order of $10^{-2}$ seconds per topology \hll{on Intel Xeon CPU E5-1620 @ 3.50GHz}) compared with that through a TO solver (in the order of $10^{2}$ to $10^3$ seconds per topology).}
\hll{It is also worth mentioning that the run-time cost of the model does not include the data acquisition cost for training the model, and the above comparison only makes sense for cases where the expected amount of computation (for all TO problems to be solved) is much larger than what is required for training the network. We acknowledge, nonetheless, that quantifying the total computational cost of a design task is challenging (e.g., due to unknown designer preferences), and theoretical upper bounds on the data size for neural networks are yet to be established. However, addressing these issues is beyond the scope of this paper.} 
\section{Discussion}
\label{sec:discussion}
We now discuss the remaining issues related to learning manifolds of optimal solutions using neural networks.

\subsection{Curse of dimensionality and potential solutions}
\highlight{So far we assumed that the pre-defined network architecture is able to capture the underlying manifold in high-dimensional spaces. As we see from the two cases, increasing the latent dimension of the manifold from one to three already raised the data demand significantly, indicating that the manifold of optimal topologies underlying Case 2 is much more ``complex'' than that of Case 1. Indeed, from Narayanan and Mitter~\cite{narayanan2010sample}, the sampling complexity of learning a manifold (to within a specified tolerance) is exponential on the intrinsic dimension (the dimension of the input), linear on the intrinsic volume (the size of the input space), and polynomial on the curvature of the manifold.}

\highlight{This leads to two legitimate concerns. The first regards model sufficiency: pre-defining the network architecture of the generator could be a stab in the dark when a new manifold is to be learned, as we do not know whether the network is sufficiently flexible to fit to the manifold. The second regards data sufficiency: real-world design problems may have solution manifolds that are too data-demanding for active learning alone to handle.}

Lei et al.~\cite{lei2018geometric} recently proposed an approach to the first concern in the context of piecewise linear networks (with ReLU activations). It is shown that both the manifold and the network complexity can be measured by the number of polyhedral cells induced respectively by the geometry of the manifold and the architecture of the network. Further studies following these complexity measures may lead to protocols for determining the network architecture of the generator before learning of generator weights takes place, essentially by computing an upper bound on the number of planes needed to locally and linearly approximate the underlying manifold. An efficient algorithm for doing so, however, is yet to be developed.

\highlight{The second concern, however, is more critical to the application of the proposed method. One potential solution is based on the insight that the governing equations for training can be derived at arbitrary spatial resolution of the structure. While the intrinsic dimension and volume do not change across resolutions, we hypothesize that lowering the resolution will reduce the manifold curvature, and thus reduce the sampling complexity. In the context of TO, with lower resolution of the structure (i.e., less elements), we expect an easier learning problem. Based on this hypothesis, it is possible that a hierarchical network architecture can be learned progressively to alleviate the curse of dimensionality: At each spatial resolution level, we learn a generator that predicts the transition from coarse solutions to the ones that satisfy the optimality conditions at this level. The coarse solutions are proposed by the generator learned from the lower-level resolution. The investigation of this approach will be reported in a separate paper.} \highlight{It should be reiterated that despite the inevitable scalability challenge, the value of learning models through domain-specific theories is clearly demonstrated in this study.}

\subsection{From theory-driven data selection to full theory-driven learning}
\begin{figure}[htp]
\centering
\includegraphics[width=1.0\textwidth]{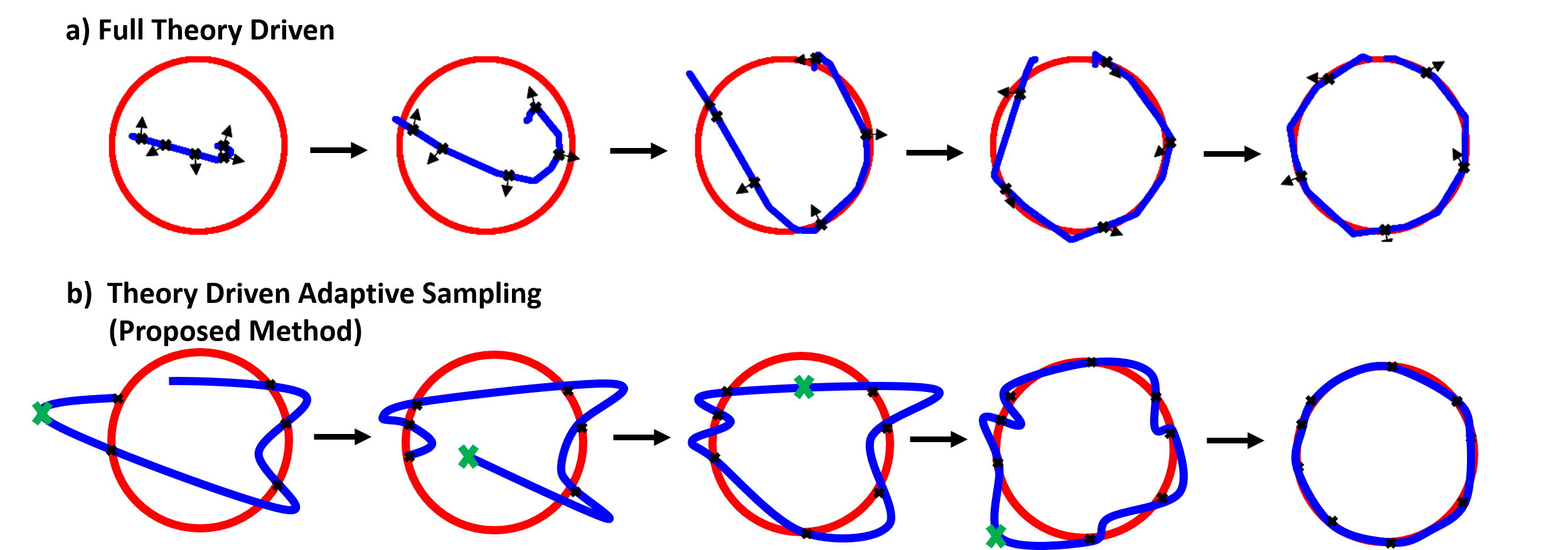}
\caption{(a) Full theory-driven learning: In each iteration, random inputs are picked to compute the improvement directions of the model towards the ground truth. (b) Theory-driven adaptive sampling (this paper): Sample inputs are selected based on their violation to the optimality conditions. The corresponding ground truth (the crosses) are revealed, which informs the improvement directions of the model. Best viewed online.}
\label{fig:circle}
\end{figure}

Another direction to explore is based on the note that the proposed method does not solve the learning problem \eqref{eq:prob} directly. Rather, we collect true solutions and fit a generator (i.e., a neural network) to it, with the hope that by intelligently collecting true solutions on fly, the fit model will effectively converge to the true solution manifold governed by the optimality conditions. An interesting question is whether solving \eqref{eq:prob} directly through a gradient-based method can be achieved and will be more effective than the presented method. If we consider the presented learning mechanism as theory-driven data selection, then solving \eqref{eq:prob} directly will be full theory-driven learning. More concretely, in each iteration of the learning, we would need to solve a batch of TO problems \textit{partially}, i.e., finding feasible topologies that reduce the violation to the optimality conditions, and use the resultant changes in topologies to update the solution generator. The key difference between the presented method and a full theory-driven learning mechanism is that while the former guarantees optimality of known solutions as long as the network is flexible enough to fit through these solutions, the latter does not have such a guarantee at any time during the training; instead, it requires less cost per iteration (since it does not completely solve TO problems), and may afford more iterations (batches of TO problems). Fig.~\ref{fig:circle} visualizes the difference between the two using a simple 2D illustration, where the circle is the unknown solution manifold, the curve represents a solution generator, and the dots are sampled solutions.

\subsection{Interpretability of the generators}
One additional challenge we face is the interpretation of the learned generator. Given the fact that the generators perform reasonably well in the two cases, one would like to visualize what they learn, e.g., local structural features that serve as puzzle pieces and lead to low-compliance structures when assembled. However, an investigation into the learned generators shows that such interpretable knowledge are not evident from the visualization of the network parameters. One potential reason is that the arbitrarily chosen network architecture may introduce confounding hidden nodes that decompose interpretable features that could have existed. 
One solution could be to impose regularization on all network parameters to be learned. 
Yet this will introduce more hyper-parameter tuning as a result.

\subsection{Value of learning manifolds of optimal designs to designers}
Last but not least, we shall come back to discuss and reiterate why and when learning manifolds of optimal designs has a value to designers. As we discussed in the introduction, there are cases where a large number of topology optimization problems (or other forms of optimization problems) need to be solved with only parametric differences. Such cases include when the TO is nested in a larger-scale system-level optimization, or when problem settings of the TO are required to be explored by human designers. In addition, designers may solve problems without knowing that similar ones have been previously solved by others. The proposed learning mechanism would not only allow a machine to accumulate and learn from solutions to similar problems, but also \textit{effectively} practice by itself to reinforce its intuition at quickly solving common sets of problems, and thus may reduce the computational and energy cost of design tasks of growing complexity. 

On the other hand, the learning itself requires solving more problems during adaptive sampling. Hence, it would be ideal to understand whether the expected total cost of solving a set of problems in the future surpasses the potential cost of learning. If so, the learning mechanism would have a value. The comparison between these two costs, however, will not be straight forward, as we do not know the learning cost beforehand for a certain performance threshold of the resultant generator. Therefore, a performance bound of the generator along the sample size will need to be developed to guide the decision on whether the learning has a practical value or not. 

\section{Conclusion}
\label{sec:conclusion}
We were motivated by the lack of knowledge accumulation capability of existing computational solvers for design problems. This drawback of machines hampers them from quickly creating good solutions in response to changing requirements in real-world design processes. Our solution to this end was to create a solution generator that adaptively learns from true solutions from a distribution of problems and predicts solutions to other unseen problems from the same distribution. The generator was modeled by a feedforward neural network, and thus produces solutions in one-shot, as opposed to through iterations as in conventional approaches. Our key contribution was the introduction of problem-specific optimality conditions as a tractable validation measure to enable more effective learning. We highlight that computing the violation of a generated solution to the optimality conditions requires only a single finite element analysis, while comparing the generated solution with the ground truth would require an entire topology optimization process, which requires thousands of finite element analysis for the problems we studied. 
We showed through two case studies that the proposed learning algorithm achieves significantly better generalization performance than the benchmarks under the same computational cost. \hll{While more sophisticated topology optimization settings should be tested, the proposed method is generally applicable to tasks of learning distributions of optimal solutions, provided that the optimality conditions can be derived and computed at low cost.} Source codes for reproducing results from the paper are available \href{https//github.com/DesignInformaticsLab/Theory_Driven_TO}{here}.

\section*{Appendix}

\subsection*{Topology optimization through augmented Lagrangian}
\label{sec:al}
The topology optimization problem of \eqref{eq:top} is solved using Alg.~\ref{alg:al}. Note that other algorithms, such as the method of moving asymptotes~\cite{svanberg1987method}, are also applicable.
\begin{algorithm}
\label{alg:al}
\SetKwInOut{Input}{input}
\SetKwInOut{Output}{output}
\Input{Problem parameters $\alpha$, $\beta$, $p$, $r_e$, $R_e$, $\beta_t$, $\textbf{s}$}
\Output{Design variable $\textbf{x}$}
Set algorithmic parameters $\epsilon_{al} = 1$, $\epsilon_{opt} = 10^{-3}$, initial guess $\textbf{x} = \alpha \textbf{1}$; \par
Pre-compute neighbourhood $\mathcal{M}_e$ and $\mathcal{N}_e$, and filter weights $\omega_{i,j}$;\par
\While{$\beta < \beta_t$}{ 
    \tcc{gradually change the problem}
    Set AL parameters: $r_0=1$, $r_1 =1$, $\mu_0 = \mu_1 = 0$, $\eta_0 = \eta_1 = 0.1$, $\epsilon = 1$, $\delta \textbf{x} = 10^6\textbf{1}$; \par
    Compute $\tilde{\textbf{x}}$, $\boldsymbol{\rho}$, $\bar{\boldsymbol{\rho}}$, $\textbf{K}$, $\textbf{u}$, $f$, $g_0$, and $g_1$; \par
    \While{$\max |\delta \textbf{x}| > \epsilon_{al}$ or $g_0 > 0$ or $g_1 > 0$}{
        \tcc{solve the constrained problem}
        Set $\delta \textbf{x} = 10^6\textbf{1}$; \par
        \While{$\max |\delta \textbf{x}| > \epsilon$}{
            \tcc{solve the unconstrained problem}
            Set learning rate $a = 10^{-3}$; \par
            Compute $\nabla_{\textbf{x}}f$, $\nabla_{\textbf{x}}g_0$, $\nabla_{\textbf{x}}g_1$, and $\delta \textbf{x} = \nabla_{\textbf{x}}f + (\mu_0 + 2g_0/r_0)\nabla_{\textbf{x}}g_0 (g_0>0) + (\mu_1 + 2g_1/r_1)\nabla_{\textbf{x}}g_1 (g_1>0)$; \par
            \While{1}{
                \tcc{line search}
                Set $\Delta \textbf{x} = -a \delta \textbf{x}$, clip each element of $\Delta x$ to $[-0.1, 0.1]$;\par
                Set $\textbf{x}' = \textbf{x} + \Delta \textbf{x}$;\par
                Compute $\boldsymbol{\rho}'$, $\bar{\boldsymbol{\rho}}'$, $\textbf{K}'$, $\textbf{u}'$, $f'$, $g_0'$, and $g_1'$ based on $\textbf{x}'$; \par
                Compute $L = f + \mu_0g_0 + 0.5g_0^2/r_0 + \mu_1g_1 + 0.5g_1^2/r_1$; \par
                Compute $L' = f' + \mu_0g_0' + 0.5g_0'^2/r_0 + \mu_1g_1' + 0.5g_1'^2/r_1$; \par
                \uIf{$L'-L>0$}{
                    \tcc{if learning rate is too high}
                    Set $a = 0.5a$; \par
                    }
                \Else{
                    $\textbf{x} = \textbf{x}'$;\par
                    }
                }
            }
        \tcc{update augmented Lagrangian parameters}
        \For{$i=0,1$}{
            \uIf{$g_i < \eta_i$}{
                Set $\mu_i = \mu_i + 2g_i/r_i$, $\eta_i = 0.5\eta_i$; \par
                }
            \Else{
                Set $r_i = 0.5r_i$; \par 
                }
        }
            
        }
    Set $\beta = 2\beta$; \par
    }
\caption{Topology optimization through Augmented Lagrangian (AL)}
\end{algorithm}

\bibliographystyle{elsarticle-num}
\bibliography{ref}

\end{document}